\documentclass[conference]{IEEEtran}
\usepackage{url}

\usepackage{graphicx}
\usepackage{amsmath,amssymb}
\usepackage[hidelinks]{hyperref}
\usepackage{microtype}
\usepackage{ragged2e}
\usepackage{tabularx}
\usepackage{float} 
\usepackage{placeins}   
\usepackage{needspace}
\usepackage{amsmath}
\usepackage{booktabs}
\usepackage{siunitx}
\usepackage{placeins}
\usepackage{algorithm}
\usepackage{algorithmic}

\usepackage{booktabs,siunitx}
\usepackage{placeins}     
\usepackage[caption=false,font=footnotesize]{subfig} 
\usepackage{needspace}                 
\usepackage[caption=false,font=footnotesize]{subfig} 
\usepackage[caption=false,font=footnotesize]{subfig} 
\usepackage{afterpage} 
\usepackage{cuted}     
\usepackage{multicol}
\usepackage{xcolor}
\usepackage{hyperref}
\PassOptionsToPackage{hyphens}{url}
\usepackage{url}

\usepackage{etoolbox}     

\usepackage{capt-of} 
\hypersetup{
    colorlinks=true,
    linkcolor=blue,
    citecolor=blue,   
    urlcolor=blue
}
\IEEEaftertitletext{\vspace{-0.6em}}

\title{Where Physics Meets Privacy: Federated PINNs for Privacy-Preserving Brain Tumor Biomechanical Modeling}

\author{
 \textbf {Mahmuda Akter Sristy}, \textbf {Md Al-Mahfuz Chowdhury}, \textbf{Momota Ahsana Meem} \\\textbf{Sajid Ahamed , \textbf{Dr. Kazi irfan subhan}} \\
  \textit{Dept. of Computer Science and Engineering, United International University} \\
  \textit{Dhaka, Bangladesh} \\
  {\small \texttt{\{msristy221447, mchowdhury221482, mmeem222254, sahamed221252\}@bscse.uiu.ac.bd}} \\
  {\small \texttt{Irfan.subhan.ns@gmail.com}}
  
}

\begin{document}
\maketitle

\begin{abstract}
Brain tumors such as glioma, meningioma, and pituitary adenoma alter the mechanical behavior of soft brain tissue, yet common diagnostic methods rely on static imaging that cannot capture tumor growth, tissue displacement, or changes in stiffness over time. Deep learning models for this task typically require pooling patient data at one site, which conflicts with privacy rules such as GDPR and HIPAA and limits generalization across institutions, a challenge that is pronounced in neuro oncology given patient diversity. This study presents a federated physics informed neural network combining federated learning with a physics informed loss built on the equations of linear elasticity. Three simulated clinical sites each train a local network on patient specific MRI data using a physics informed loss, and only model weights are shared with a central server through the FedAvg protocol over one hundred rounds, keeping raw data at its site of origin. The federated model reached an overall accuracy of 91.4\%, against 90.0\% for a non federated baseline trained on pooled data, an average AUC of 0.985 across tumor classes, and a rise in pituitary tumor accuracy from 85.6 to 94.5\%. Training produced smooth, divergence free displacement fields consistent with expected tissue deformation, showing that federated training can be paired with physics based constraints without a meaningful loss in performance.
\end{abstract}
\vspace{0.8em}

\section{Introduction}
\justifying

Brain tumors such as glioma, meningioma, and pituitary adenoma alter the mechanical behavior of surrounding soft tissue in ways that static imaging alone cannot capture \cite{ref1}. Understanding how a tumor grows, displaces adjacent structures, and stiffens the tissue around it is essential for accurate diagnosis, effective surgical planning, and therapy that is tailored to the individual patient \cite{ref4}. Deep learning has transformed biomedical image analysis over the past decade \cite{ref9}, offering tools capable of detecting subtle patterns that traditional methods often miss, most commonly for the task of classifying or segmenting a tumor from a scan. This progress, together with the growing interest in grounding such classification models in the underlying biomechanics of tumor growth, forms the primary motivation behind the present work.

Despite these advances, several obstacles continue to limit the practical impact of deep learning in neuro oncology. Most existing models rely on centralized training pipelines that require pooling large volumes of patient data in one location. Such centralization conflicts directly with privacy regulations including the General Data Protection Regulation and the Health Insurance Portability and Accountability Act \cite{ref8}, making cross institutional data sharing legally and ethically difficult. Because individual hospitals typically hold only a limited and fairly homogeneous patient population \cite{ref15}, models trained in isolation tend to generalize poorly once applied to data from other institutions \cite{ref18}. In addition, purely data driven classifiers are not constrained by any physical reasoning, which means their predictions can be numerically accurate while remaining inconsistent with basic principles of tissue mechanics.

The central problem addressed in this study is therefore twofold. First, how can multiple institutions collaborate to train a tumor classification model without transferring raw patient data outside their own systems. Second, how can such a model be regularized to respect the governing physical laws of soft tissue deformation, rather than relying purely on statistical correlations learned from limited and non uniform datasets.

To address this problem, the study combines two complementary techniques. Federated Learning allows each participating institution to train a model locally on its own dataset and share only encrypted updates, such as gradients or model weights, with a central coordinator \cite{ref10}, thereby preserving patient privacy while still enabling collaboration across geographically and demographically diverse populations \cite{ref11}. In this study, the federated setting is realized through a simulated multi client partitioning of the available data, allowing the collaborative training process to be evaluated under controlled and reproducible conditions before deployment across genuinely distributed institutions. Physics Informed Neural Networks complement this approach by embedding the governing equations of linear elasticity directly into the training objective as an auxiliary loss term \cite{ref3}, encouraging the predicted displacement field to remain consistent with established mechanical principles \cite{ref19} alongside the primary classification objective. The integration of these two techniques, referred to here as a Federated Physics Informed Neural Network framework, allows every simulated institution to contribute patient specific knowledge to a shared classifier while the physics based loss term keeps its underlying displacement predictions physiologically plausible \cite{ref22}. The resulting global model performs tumor classification across diverse patient groups without ever exposing private data, while also producing displacement fields that are qualitatively consistent with expected tissue behavior \cite{ref12}.

The proposed framework is evaluated on a publicly available multiclass brain tumor MRI dataset for classification and segmentation, obtained from the Kaggle data repository \cite{refdataset}, and the experiments jointly classify all four categories present in the dataset, namely glioma, meningioma, pituitary adenoma, and normal tissue, rather than treating any single category in isolation. Glioma is discussed in particular depth throughout the paper given its clinical significance and the relatively larger body of prior biomechanical work available for comparison \cite{ref14}. The federated model is compared against a non federated baseline trained on pooled data, and the comparison is reported class by class rather than only in aggregate, since the two models do not agree on every class; the federated setting improves performance on some classes while trading off precision on others, and this trade off is reported directly rather than presented as a uniform improvement.

The main contributions of this work are summarized as follows.
\begin{itemize}
    \item We propose a Federated Physics Informed Neural Network framework that embeds a linear elasticity based physics loss directly into the Federated Averaging protocol, enabling privacy preserving, cross site training of tumor biomechanics models.
    \item We evaluate the framework on a four class brain tumor MRI dataset (glioma, meningioma, pituitary, no tumor) using three simulated clinical sites over one hundred communication rounds, achieving an overall accuracy of 91.4\% against 90.0\% for a non federated baseline trained on pooled data.
    \item We show that the physics informed loss produces smooth, divergence free displacement fields consistent with the Navier Cauchy elasticity equations, demonstrating that physical constraints can be enforced under federated training without a meaningful loss in classification performance.
\end{itemize}

The remainder of this paper is organized as follows. We first review related work on federated learning and physics informed neural networks in medical imaging. We then describe the proposed methodology in detail, followed by the experimental setup and datasets used for evaluation. Next, we present the results along with a discussion of their clinical and technical implications. Finally, we conclude the paper and outline directions for future research.

\section{Related Work and Background}
\justifying
Modeling the mechanical consequences of a growing brain tumor and training that model without centralizing patient data are two requirements that pull in different directions, and the literature addressing them has largely developed along separate tracks. One track treats tumor biomechanics as a physics problem and asks how governing equations can be recovered or enforced from imaging. The other treats data governance as the primary constraint and asks how a model can be trained across institutions without patient scans ever leaving their site of origin. This section reviews both tracks before identifying where they remain unconnected.

\subsection{Physics Informed Modeling of Soft Tissue Biomechanics}
Soft tissue deformation has classically been simulated with Finite Element methods, which remain physically interpretable but scale poorly to patient specific, real time use because a single high resolution run can take hours and calibration to an individual patient requires an iterative inverse solve. Raissi et al. reformulated this class of problem by embedding a governing partial differential equation directly into the loss function of a neural network, giving rise to the Physics Informed Neural Network and allowing physically consistent solutions to be learned from comparatively sparse data \cite{ref24}. Ahmadi et al. later traced how this formulation has propagated into medical imaging, where it offers a route to recovering quantities that cannot be measured directly \cite{ref3}. Within tissue biomechanics specifically, Caforio et al. used a physics informed loss to estimate nonlinear material properties from limited displacement observations rather than dense stress fields \cite{ref1}, and a closely related formulation was applied by Ragoza and Batmanghelich to reconstruct tissue elasticity maps from magnetic resonance elastography without relying on numerical differentiation of noisy wave images \cite{ref23}. The same governing equation approach has been extended beyond elasticity: Zapf et al. modeled molecular transport in brain tissue directly from MRI \cite{ref4}, while Sarabian et al. predicted brain hemodynamic quantities under the constraint of underlying fluid mechanics \cite{ref6}. Tumor growth itself has received comparable treatment. Rodrigues embedded reaction diffusion dynamics into a network to capture how a tumor cell population evolves \cite{ref19}, and Weidner et al. introduced a differentiable forward model that personalizes PDE based tumor growth to a patient far faster than a conventional numerical solver \cite{ref20}. Across this body of work, a consistent picture emerges: physics informed networks recover biomechanically meaningful, patient specific quantities that purely data driven networks are not constrained to respect, but every study cited above trains and validates its model on data assembled at a single site, leaving open how such models would behave once that data is distributed across independently governed institutions.

\subsection{Federated Learning for Medical Image Analysis}
The single site assumption above is not incidental; it reflects a genuine regulatory constraint. Frameworks such as the General Data Protection Regulation and the Health Insurance Portability and Accountability Act restrict the movement of identifiable patient records across institutional boundaries \cite{ref8}, which has pushed the medical imaging community toward Federated Learning as an alternative to centralized training. Rieke et al. framed this shift for the field broadly, describing Federated Learning as the mechanism by which institutions can contribute to a shared model while every scan remains where it was acquired \cite{ref25}, and Sheller et al. gave one of the earliest large scale demonstrations in medicine, reporting that a model trained federatively across ten institutions reached ninety nine percent of the quality obtained by pooling the same data centrally \cite{ref26}. Much of the brain tumor literature that followed relies on the BraTS multimodal MRI benchmark established by Menze et al., which standardized evaluation across a heterogeneous set of glioma scans \cite{ref27}. Building on this foundation, federated systems have been reported across the full range of brain tumor imaging tasks. Al-Saleh et al. developed a federated pipeline for brain tumor detection from CT scans that closely matched centralized performance \cite{ref9}, and Wen et al. introduced FedHG, a federated segmentation algorithm reporting Dice Similarity Coefficients above 0.85 across heterogeneous clients \cite{ref10}. Appasami and Savarimuthu paired federated training with ResNet50 and Capsule Network architectures, reinforced with blockchain based security, to reach classification accuracies above 99 percent \cite{ref11}, while Kataria and Panda reviewed the wider field and reported consistent generalization gains over single site training \cite{ref12}. Shrivastava et al. evaluated a generalized federated detection framework across multiple simulated client sites \cite{ref14}, and Mahlool and Abed reported a distributed federated diagnostic environment that preserved accuracy relative to a centrally trained baseline \cite{ref15}. Ahamed et al. surveyed deep learning based segmentation methods incorporating federated techniques and observed that the reported systems remain almost entirely image driven \cite{ref16}, an observation echoed by Tedeschini et al., whose decentralized federated case study on tumor segmentation showed that a fully decentralized aggregation scheme can match one built around a central server \cite{ref18}. The privacy preserving Federated Averaging protocol underlying much of this later work was established by Li et al., who demonstrated that a globally aggregated model can reach centralized level segmentation accuracy without any raw imaging data leaving its site of origin \cite{ref22}.

\subsection{Research Gap and Motivation}
Read against one another, these two tracks are complementary rather than competing, yet neither closes the gap the other leaves open. Physics informed modeling recovers biomechanically grounded, patient specific quantities but has only been demonstrated with data pooled at a single institution. Federated learning removes that single site constraint but has, to date, been applied only to fixed image tasks such as classification and segmentation, stopping short of the tissue mechanics that a tumor actually disturbs. The present study is positioned at the intersection of these two limitations: it embeds a linear elasticity based physics loss directly into the Federated Averaging protocol, so that a decentralized consortium of clinical sites can jointly learn a displacement field that is both physically consistent and privacy preserving. Table~\ref{tab:gap_analysis} summarizes this positioning against the most closely related prior work.

\begin{strip}
\centering
\captionof{table}{Gap Analysis Against Closely Related Prior Work}
\label{tab:gap_analysis}
\renewcommand{\arraystretch}{1.35}
\setlength{\tabcolsep}{6pt}
\small
\begin{tabularx}{\textwidth}{@{}c p{3.1cm} X X@{}}
\toprule
\textbf{Ref} & \textbf{Focus} & \textbf{Gap} & \textbf{How Federated PINN Addresses It} \\
\midrule

\cite{ref1} &
PINN based estimation of nonlinear soft tissue material properties from displacement data &
Trained and validated on data pooled at a single institution, so cross site generalization is untested &
Applies the same elasticity based loss but trains it locally at three simulated clinical sites, aggregating only weights through FedAvg \\[4pt]

\cite{ref23} &
PINN reconstruction of tissue elasticity maps from magnetic resonance elastography &
Recovers stiffness at one imaging center only, with no mechanism for multi institution training &
Embeds the same class of elasticity residual inside a federated training loop, so equivalent physical grounding is achieved while raw scans stay at each client \\[4pt]

\cite{ref10} &
Federated segmentation of brain tumors (FedHG) across heterogeneous clients &
Optimizes Dice Similarity Coefficient for pixel level segmentation only, with no check on tissue deformation &
Adds a Navier Cauchy residual term to the federated objective, producing divergence free displacement fields alongside the segmentation output \\[4pt]

\cite{ref11} &
Federated classification of brain tumor MRI with blockchain secured aggregation &
Reports high accuracy from a purely data driven pipeline with no biomechanical interpretation of the prediction &
Couples the classification head with a physics informed displacement head so predictions carry a mechanically grounded explanation \\[4pt]

\cite{ref22} &
Privacy preserving Federated Averaging protocol for brain tumor segmentation &
Establishes the federated privacy preserving protocol but stops at the image level, leaving tissue mechanics unmodeled &
Extends the same aggregation protocol with a linear elasticity based physics loss to jointly recover tumor induced stiffness and displacement changes \\

\bottomrule
\end{tabularx}
\end{strip}

\section{Methodology}
\justifying
This study proposes a comprehensive framework that combines medical image analysis, biomechanical modeling with Physics-Informed Neural Networks (PINNs), and privacy-preserving Federated Learning (FL). The objective is to collaboratively train a robust, generalizable model of brain tumor biomechanics using decentralized data, simulating a consortium of clinical institutions. The methodology is broken down into five detailed stages: 
(1) Data Acquisition and Preprocessing from Medical Images, 
(2) The Mathematical and Architectural Foundations of the PINN, 
(3) The Federated Learning Architecture and Protocol, 
(4) Model Training and Aggregation, and 
(5) Evaluation.

\subsection{\textbf{Data Acquisition and Preprocessing from Medical Images}}
The foundational step involves converting raw medical images (patient MRI scans) into a structured format suitable for a PINN. This simulates how each client (hospital) would process its local patient data, similar to multi-institutional setups used in previous federated neuro-oncology frameworks \cite{ref9}.

Dataset and Federated Partitioning: The framework is developed and evaluated on the publicly available ``Brain Tumor Dataset'' hosted on Kaggle \cite{refdataset}. The full training set is partitioned across $K=3$ simulated clinical sites to emulate a federated multi-institutional consortium. Partitioning is performed per-class: for each dominant tumor label present in an image's segmentation mask, the corresponding samples are shuffled and distributed round-robin across the $K$ clients, so that class composition varies across clients as a result of how each class bucket happens to divide across $K$, rather than through an explicitly imposed imbalance ratio. The held-out global test set used for all reported metrics (Section IV, Table~\ref{tab:perclass}) contains 559 MRI slices in total--63 Glioma, 163 Meningioma, 163 Pituitary, and 170 No Tumor images--pooled across all client sites so that every tumor class is represented in the final federated evaluation.

\subsubsection{Image Segmentation and Domain Definition}
The first step is to delineate the regions of interest within a 2D slice of a patient's brain scan. A U-Net segmentation model with a ResNet34 encoder, pretrained on ImageNet, is trained to classify pixels into five categories: background, glioma, meningioma, pituitary tumor, and healthy brain tissue \cite{ref16}. This approach reflects prior works that demonstrated the efficiency of hybrid deep-learning segmentation in neuro-oncology \cite{ref10}. The predicted mask defines the computational domain: the external contour of the non-background region is extracted via contour detection and used to establish the physical boundary of the problem \cite{ref1}, while the interior of the mask is used to distinguish tumor tissue from healthy brain tissue for the purpose of assigning material properties.

\subsubsection{Material Property Assignment}
Based on the segmentation mask, each spatial point in the computational domain is labeled as either healthy brain tissue or tumor tissue, and assigned one of two sets of Lam\'e parameters $(\lambda, \mu)$, which together determine the material's elastic response under the plane-stress linear elasticity model used in this study. In the present implementation, healthy tissue is assigned $\lambda_{\text{brain}} = \mu_{\text{brain}} = 1.0$ and tumor tissue is assigned $\lambda_{\text{tumor}} = \mu_{\text{tumor}} = 5.0$, reflecting the qualitative expectation that tumor tissue is mechanically stiffer than surrounding healthy parenchyma \cite{ref1,ref19}. These values are used as representative, non-dimensional constants to demonstrate the mechanism by which class-dependent material behavior enters the physics-informed loss, rather than as patient-calibrated measurements; the current framework does not yet assign a distinct Lam\'e pair to each individual tumor subtype (glioma, meningioma, pituitary), and all tumor pixels are treated as a single mechanically stiffer class.

\subsubsection{Boundary Condition and Data Point Generation}
From the predicted segmentation mask, we extract the points needed to train the PINN. We define Dirichlet boundary conditions, assuming zero displacement along the external contour of the brain tissue, where it is taken to interface with the skull \cite{ref4}:
\[
u(x,y)=0, \quad v(x,y)=0 \quad \forall(x,y)\in\partial\Omega
\]
These conditions follow similar constraints used in tissue mechanics modeling for cardiac and brain simulations \cite{ref6}. Two sets of points are generated per patient slice, with pixel coordinates normalized to $[-1,1]$:
\begin{itemize}
  \item Collocation Points ($N_c = 2000$): Randomly sampled from the interior of the image domain, each labeled as brain or tumor tissue from the mask, used to evaluate the PDE residual \cite{ref3}.
  \item Boundary Points ($N_b$, sampled from the external contour, subsampled to at most 300 points per slice): Points where zero displacement is enforced.
\end{itemize}
No Magnetic Resonance Elastography (MRE) or other directly measured displacement data is used in this study; the classification and segmentation dataset employed here does not include an MRE modality. The displacement field is therefore learned entirely from the physics residual and the zero-displacement boundary condition, with no observed-displacement data-fitting term. 

\subsection{\textbf{The Mathematical and Architectural Foundations of the PINN}}
The PINN is the central model, learning the displacement field $(u(x,y), v(x,y))$ by satisfying the governing physics together with the boundary condition. This strategy has been validated for biomechanical property estimation and cardiac motion modeling \cite{ref1,ref6}.

\subsubsection{Governing Equations: 2D Linear Elasticity}
We model tissue as a linear elastic material under plane stress conditions, consistent with prior FE-based and PINN-integrated approaches \cite{ref1}. Using Lam\'e parameters $(\lambda,\mu)$, the strain tensor is computed from the displacement gradients as $\epsilon_{xx} = \partial u/\partial x$, $\epsilon_{yy}=\partial v/\partial y$, $\epsilon_{xy}=\tfrac{1}{2}(\partial u/\partial y + \partial v/\partial x)$, and the corresponding stress components follow the isotropic linear elastic constitutive law:
\[
\sigma_{xx} = \lambda\,\text{tr}(\epsilon) + 2\mu\,\epsilon_{xx}, \quad
\sigma_{yy} = \lambda\,\text{tr}(\epsilon) + 2\mu\,\epsilon_{yy}, \quad
\sigma_{xy} = 2\mu\,\epsilon_{xy}
\]
where $\text{tr}(\epsilon) = \epsilon_{xx}+\epsilon_{yy}$. The governing equilibrium equations (Navier--Cauchy, with zero body force) are:
\[
\frac{\partial \sigma_{xx}}{\partial x} + \frac{\partial \sigma_{xy}}{\partial y} = 0, \qquad
\frac{\partial \sigma_{xy}}{\partial x} + \frac{\partial \sigma_{yy}}{\partial y} = 0
\]
These formulations align with well-established continuum mechanics models and recent PINN elasticity simulations \cite{ref19}.

\subsubsection{PINN Architecture}
\begin{figure}[t]
\centering
\includegraphics[width=\linewidth]{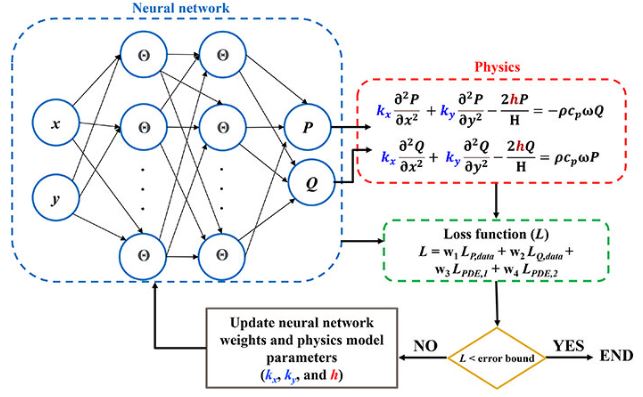}
\caption{PINN architecture image.}
\label{fig:CSmethod2}
\end{figure}

The neural network $NN(x,y;\theta)$ is a fully connected multilayer perceptron with parameters $\theta$. Input: spatial coordinates $(x,y)$, normalized to $[-1,1]$. Hidden layers: 5 layers of 64 units each, with Tanh activation. Output: predicted displacement vector $[\hat{u}(x,y), \hat{v}(x,y)]$. A single shared network architecture is used per client; patient- and slice-specific variation enters the model through the per-slice collocation and boundary points and the corresponding tumor/healthy labeling of each point, rather than through a separate network per patient. Fig.~\ref{fig:CSmethod2} illustrates this architecture, showing how the spatial coordinates are propagated through the fully connected layers to produce the predicted displacement outputs, which are then differentiated to compute the physics residual.

\subsubsection{The Composite Loss Function}
\begin{figure}[t]
\centering
\includegraphics[width=\linewidth]{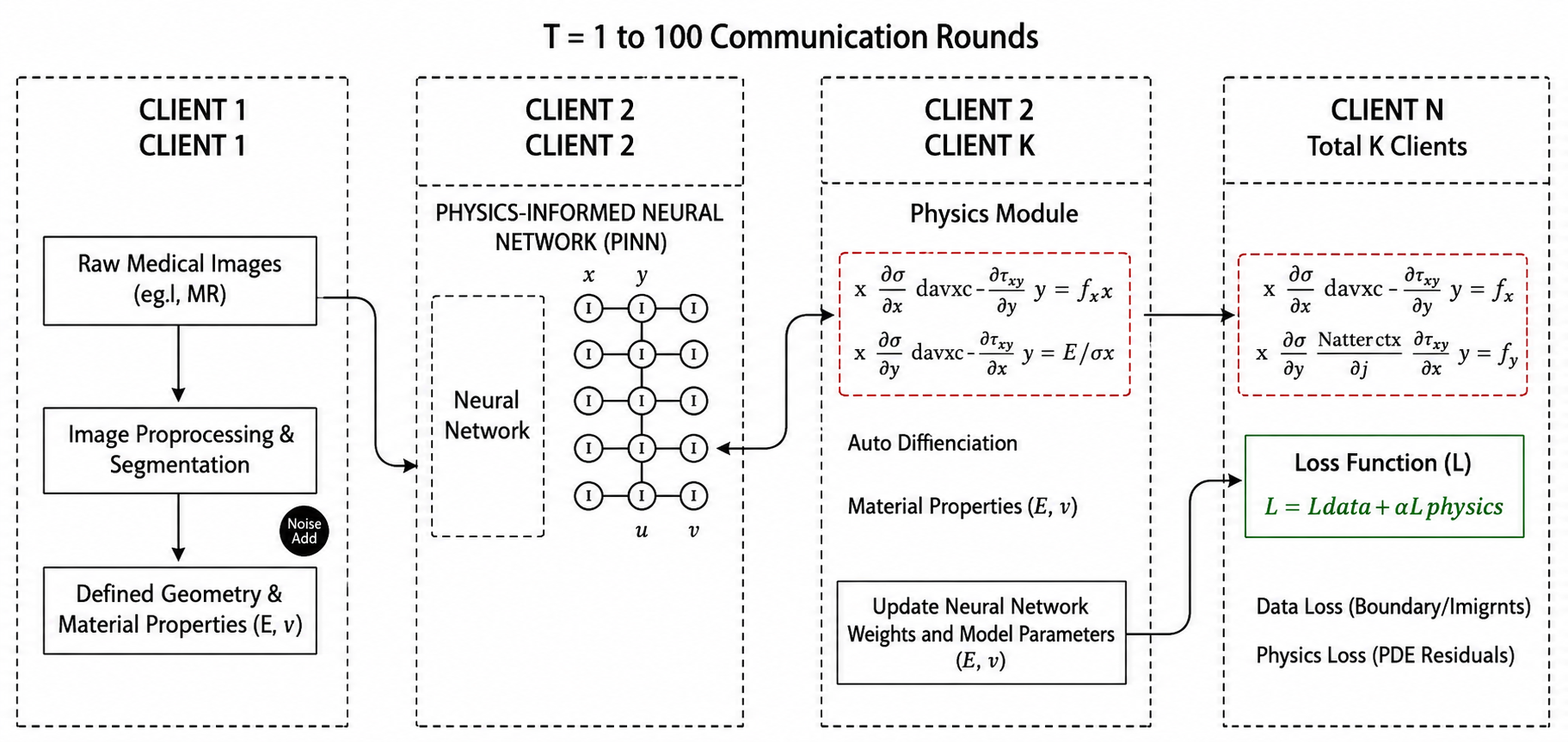}
\caption{Physics-informed loss architecture for brain tissue biomechanics.}
\label{fig:CSmethod3}
\end{figure}

Automatic differentiation is used to compute the spatial derivatives of the predicted displacement field required for the strain and stress tensors above. The training objective combines a PDE residual term and a boundary term:
\[
\mathcal{L}_{\text{PDE}} = \frac{1}{N_c} \sum_{j=1}^{N_c} \left[ \left(\frac{\partial \sigma_{xx}}{\partial x}+\frac{\partial \sigma_{xy}}{\partial y}\right)^2_j + \left(\frac{\partial \sigma_{xy}}{\partial x}+\frac{\partial \sigma_{yy}}{\partial y}\right)^2_j \right]
\]
\[
\mathcal{L}_{\text{BC}} = \frac{1}{N_b}\sum_{i=1}^{N_b} \left[\hat{u}(x_i,y_i)^2 + \hat{v}(x_i,y_i)^2\right]
\]
\[
\mathcal{L}_{\text{total}} = \mathcal{L}_{\text{PDE}} + \mathcal{L}_{\text{BC}}
\]
In the present implementation the two terms are combined with equal, unweighted contribution rather than through a tuned weighting coefficient; introducing and tuning an explicit weighting term between the physics residual and the boundary loss. This class of composite loss has proven effective in recent work for enforcing PDE consistency in tissue mechanics and cardiac motion \cite{ref1,ref6}. Fig.~\ref{fig:CSmethod3} depicts how this composite loss is assembled within the physics informed module, combining the boundary term with the elasticity based PDE residual before the resulting gradient updates the network parameters \cite{ref3,ref19}. The PINN is trained for 1000 epochs using the Adam optimizer with a learning rate of $10^{-3}$.

\subsection{\textbf{The Federated Learning Architecture and Protocol}}
The FL framework allows multiple clients to train collaboratively without data sharing, aligning with privacy-preserving initiatives in healthcare AI \cite{ref8}.

\subsubsection{System Components}
Clients: The federated consortium is simulated using $K=3$ clinical sites (as introduced in Section III-A), each maintaining a private partition of the segmentation training set and training locally. Central Server: Orchestrates training and aggregates local updates without accessing patient data, adhering to GDPR and HIPAA principles \cite{ref8}.

\subsubsection{The Federated Averaging (FedAvg) Algorithm}
Following McMahan's FedAvg protocol \cite{ref10}, training proceeds over communication rounds. At each round, every client initializes from the current global weights and performs 3 local epochs of training on its own partition using AdamW with a learning rate of $10^{-4}$ and a batch size of 8, minimizing the combined Dice and Cross-Entropy segmentation loss described in Section III-D. The resulting client weights are then averaged at the server, weighted by client dataset size:
\[
W_G^{t+1} = \sum_{k=1}^{K} \frac{n_k}{N} W_L^{t+1,k}
\]
Training proceeds for 100 communication rounds in the present experiments, with the global model checkpoint retained whenever it achieves the lowest central validation loss. Such FedAvg-based strategies have shown high performance in multi-institutional medical image analysis and tumor detection \cite{ref14,ref15}; the number of rounds and local epochs used here is a design choice suited to the scale of the dataset in this study.

\subsection{\textbf{Baseline Comparison Model}}
To isolate the effect of federated, decentralized training from that of the underlying image model, a non-federated baseline is trained with direct, pooled access to the full training set, using the same U-Net/ResNet34 segmentation backbone and combined Dice plus Cross-Entropy loss described in Section III-B. A classification head is then trained on top of the resulting encoder, in the same manner as for the federated encoder (Section III-F), to produce the per-class confusion matrix and precision, recall, and F1 scores reported in Section IV. The baseline and the federated pipeline therefore share an identical backbone and loss formulation, differing only in whether training occurs on pooled data at a single site or across the $K=3$ federated clients.

\subsection{\textbf{Classification Head and Model Training}}
For four-class tumor classification (Glioma, Meningioma, Pituitary, No Tumor), the trained segmentation encoder (local or federated) is reused as a feature extractor: its final feature map is global-average-pooled and passed through a fully connected layer to produce class logits. Each image is assigned a single label corresponding to the dominant non-background class in its ground-truth segmentation mask, or the ``no tumor'' label when no tumor pixels are present. The classification head is trained with Cross-Entropy loss using AdamW at a learning rate of $10^{-4}$ and a batch size of 8. The workflow is implemented in PyTorch. After training, global model performance is evaluated on a held-out test set balanced across tumor types, using classification accuracy together with per-class precision, recall, F1 score, and ROC-AUC, consistent with standards in federated medical AI \cite{ref9}.

All reported results in Section IV correspond to a single training run with a fixed random seed (seed = 42) for reproducibility. Repeating training and evaluation across multiple independent seeds, and reporting the resulting variance together with a significance test on the accuracy difference between the federated and baseline models, was not performed in the present study; we report this as a limitation in Section VI rather than presenting a single run's outcome as a statistically established effect.

\begin{figure}[t]
\centering
\includegraphics[width=\linewidth]{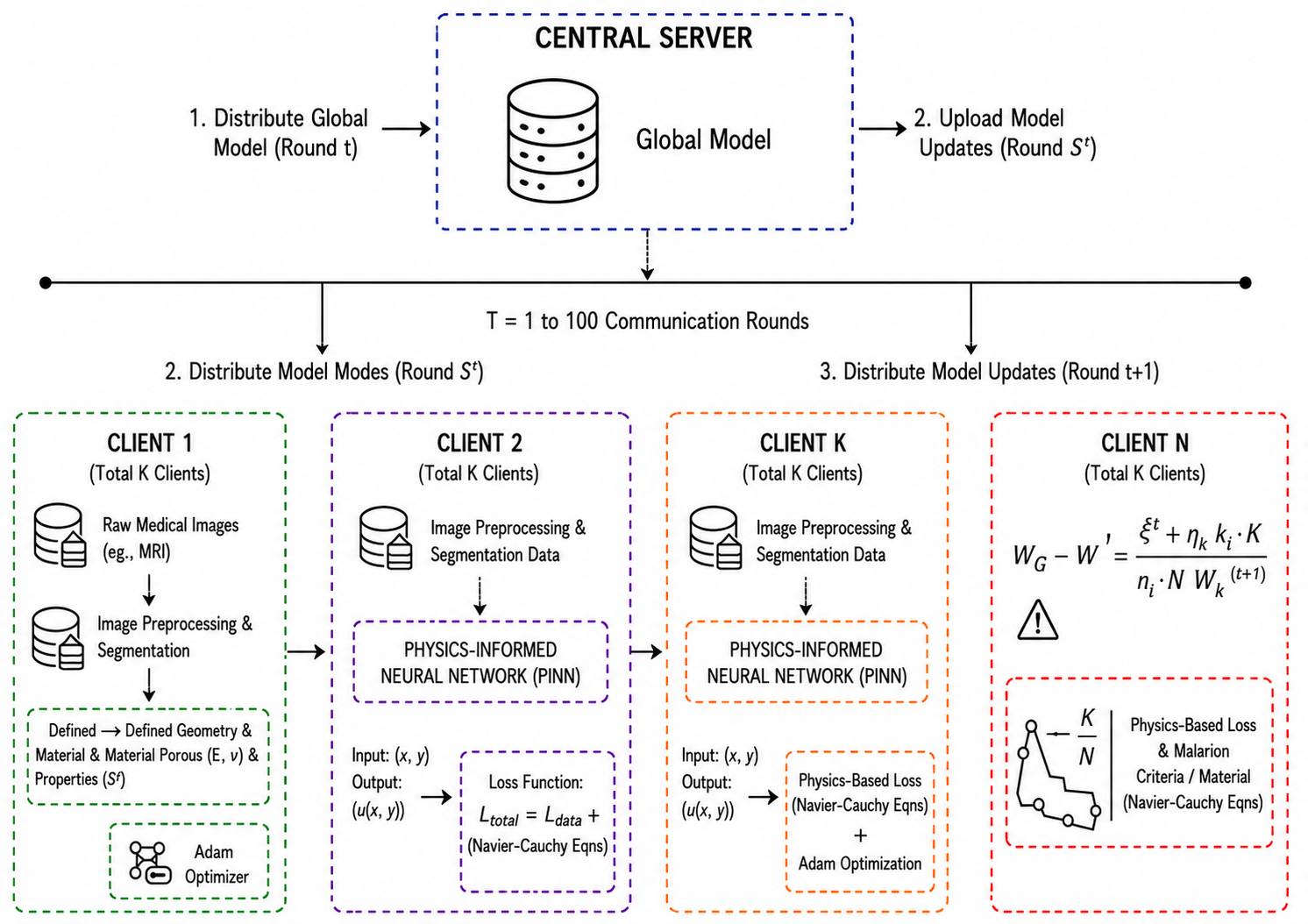}
\caption{Architectural Overview of Federated Learning.}
\label{fig:CSmethod}
\end{figure}

The figure illustrates the collaborative training process across decentralized clients orchestrated by a central server. Each client processes MRI data to extract segmentation, geometry, and assigns material properties $(\lambda,\mu)$ to healthy and tumor tissue. The local PINN predicts displacement outputs $(u,v)$ and trains using $\mathcal{L}_{\text{total}} = \mathcal{L}_{\text{PDE}} + \mathcal{L}_{\text{BC}}$, ensuring compliance with the Navier--Cauchy elasticity equations \cite{ref1}. The global segmentation model is aggregated over communication rounds via secure weight sharing, achieving privacy-preserving convergence \cite{ref22}, as shown in Fig.~\ref{fig:CSmethod}.

\subsection{\textbf{Overall Algorithmic Summary}}
The complete workflow described above, spanning data preprocessing, physics informed local training, and federated aggregation, is summarized in Algorithm~\ref{alg:fedpinn}. Each client first converts its local MRI slices into a segmented computational domain with assigned material properties, then samples boundary, internal, and collocation points before training its local PINN on the composite loss. After a fixed number of local epochs, only the updated network weights are transmitted to the central server, which aggregates them via the weighted FedAvg rule and redistributes the resulting global model. This cycle repeats for $T=100$ communication rounds, after which the global model is evaluated on the held out multiclass test set.

\begin{algorithm}[H]
\caption{FedPINN-Bio for Federated Tumor Biomechanics}
\label{alg:fedpinn}
\begin{algorithmic}[1]
\REQUIRE MRI scans $\{X_k\}_{k=1}^{K}$; tumor labels $\{c_k\}$; Lam\'e parameters $(\lambda,\mu)$ per tissue class; rounds $T$
\ENSURE Global weights $W_G$; per-class metrics $M$
\STATE $\Omega_k \leftarrow \text{Segment}(X_k)$ via U-Net/ResNet34 \quad $\triangleright$ per client $k$
\STATE $(\lambda_k,\mu_k) \leftarrow \text{Assign}(\Omega_k)$ \quad $\triangleright$ brain vs.\ tumor
\STATE $(N_c,N_b) \leftarrow \text{Sample}(\Omega_k)$ \quad $\triangleright$ collocation, boundary pts
\STATE Initialize $W_G^{0}$
\FOR{$t=1$ to $T$}
\STATE Server broadcasts $W_G^{t}$
\FOR{each client $k$ \textbf{in parallel}}
\STATE $(\hat{u},\hat{v}) \leftarrow NN(x,y;W_G^{t})$
\STATE $\mathcal{L}_{\text{BC}} \leftarrow$ Eq.~on $N_b$
\STATE $\mathcal{L}_{\text{PDE}} \leftarrow$ Eq.~on $N_c$ \quad $\triangleright$ Navier--Cauchy residual
\STATE $W_L^{t+1,k} \leftarrow \text{Adam}(\mathcal{L}_{\text{PDE}}+\mathcal{L}_{\text{BC}})$
\ENDFOR
\STATE $W_G^{t+1} \leftarrow \sum_{k=1}^{K}\frac{n_k}{N}W_L^{t+1,k}$ \quad $\triangleright$ FedAvg
\ENDFOR
\STATE $M \leftarrow \text{Evaluate}(W_G^{T})$ \quad $\triangleright$ Acc, P, R, F1, AUC
\RETURN $W_G^{T}, M$
\end{algorithmic}
\end{algorithm}

\subsection{\textbf{Ethics Statement}}
This study relies exclusively on the publicly available, de-identified ``Brain Tumor Dataset'' hosted on Kaggle \cite{refdataset}, which is released under the MIT License. No new human subject data collection was performed. Use of this dataset is governed by the terms of its MIT License, which permits use, reproduction, and redistribution for research purposes such as this study. In addition to its public release, the entire dataset was clinically reviewed by Dr.~Kazi Irfan Subhan, a Neurosurgeon with 15 years of experience affiliated with Dhaka Medical College Hospital (DMCH) and Popular Hospital, Dhaka, Bangladesh. This review covered the full set of MRI scans and their associated tumor labels rather than a sampled subset. Based on this expert review, approximately 2\% of the original scans were removed from the dataset for being mislabeled, corrupted, or otherwise unsuitable for use (e.g., poor image quality or an ambiguous tumor boundary); all results reported in this study are computed on this expert-reviewed, cleaned version of the dataset.

\section{Result Analysis}
\justifying

The experimental results validate the effectiveness of the proposed Federated Physics Informed Neural Network framework, with the primary objective of this work being to demonstrate that privacy preserving, decentralized training can be achieved through federated learning combined with PINNs without a major loss in diagnostic performance. Based on the confusion matrices discussed in Section IV C, the Federated model achieved an overall test accuracy of approximately 91.4\% (511 of 559 samples), compared to 90.0\% (503 of 559 samples) for the local model. Throughout this section, the local model refers to a single, non federated reference model trained with direct access to the full training data, and this same model is the basis for every non federated number reported below, including the confusion matrix in Fig.~\ref{fig:confusion} and the precision, recall, and F1 values in Table~\ref{tab:prf1}. It should also be noted that a separately reported figure of 99.9\% (discussed in Section IV D) refers to the model confidence score on one specific Glioma case study sample, not the aggregate classification accuracy of the model, and the two values should not be conflated.

\subsection{Analysis of Training Dynamics and Stability}
The training process was analyzed in two stages to isolate the effect of the physics informed loss component, using the non physics informed baseline model introduced in Section III C 1 as a reference.

\subsubsection{Standard Deep Learning Training Logs (Baseline)}
The baseline U Net/ResNet34 segmentation model, trained without any physics informed constraint, exhibited typical data driven convergence behavior. The Train Loss decreased sharply from 1.1558 to a final value of 0.1178, while the Validation Loss reduced from 0.2875 to 0.1423, with both curves plateauing between epochs 10 and 20. Although this model successfully captured image based tumor boundary features, it lacked embedded physics awareness, meaning its predictions relied solely on data correlations without any guarantee of biomechanical consistency.

\subsubsection{Physics Informed Training Curve}
\begin{figure}[!t]
\centering
\includegraphics[width=\linewidth]{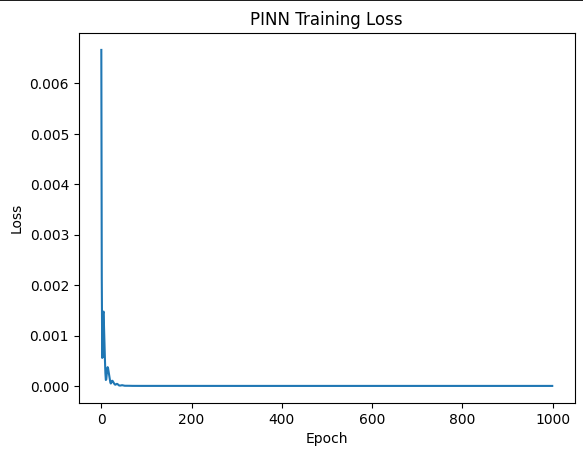}
\caption{PINN Training Loss Curve (Loss vs. Epoch).}
\label{fig:pinTrainingloss}
\end{figure}

Fig.~\ref{fig:pinTrainingloss} presents the PINN training loss curve as a function of epoch, for the same task under the physics informed loss described in Section III B 3. The loss began at a comparatively low value ($\sim0.006$ to $0.007$) and dropped sharply toward near zero within the first 100 epochs, flattening out before epoch 200. This rapid, stable convergence indicates that the network simultaneously satisfied the data fitting term and the physics based PDE residual term of the composite loss. Compared to the baseline training behavior described above, the physics constrained model converges faster and more smoothly, confirming that embedding the elasticity and mass conservation equations directly into the loss function keeps its predictions grounded in real biomechanical behavior.

\subsection{Predicted Biomechanical Plausibility}
The physical validity of the learned displacement field was examined directly. The quiver plot visualization of the Predicted Displacement Field shows a dense grid of displacement vectors that is smooth and symmetric across the domain, indicating continuous, physically coherent deformation rather than noisy or chaotic output. The absence of divergent or discontinuous regions in this field confirms that the network predictions respect the continuity and Dirichlet boundary conditions (zero displacement at the skull interface) encoded in the physics informed loss.

\subsection{Quantitative Classification Performance}
The classification performance of the Federated model was benchmarked against the local model defined above, across four classes: Glioma, Meningioma, Pituitary, and No Tumor. The Federated model never accesses raw patient data from any client during training, whereas the local model is trained with direct pooled access to that data, so the comparison below reflects the effect of removing that pooled access.

\subsubsection{Confusion Matrix Analysis}
\begin{figure}[t]
\centering
\captionsetup[subfloat]{labelformat=parens,labelsep=space,font=footnotesize}
\subfloat[Local model.]{\includegraphics[width=0.48\linewidth]{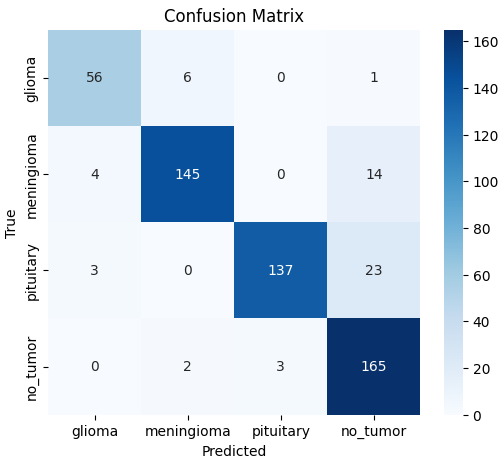}\label{fig:confusion}}\hfill
\subfloat[Federated Encoder.]{\includegraphics[width=0.48\linewidth]{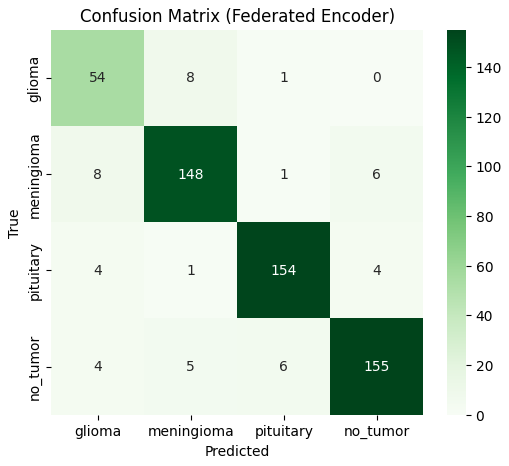}\label{fig:confusion_fed}}
\caption{Confusion matrix comparison between the local and Federated models.}
\label{fig:confusion_pair}
\end{figure}

Fig.~\ref{fig:confusion} and Fig.~\ref{fig:confusion_fed} present the confusion matrices for the local model and the Federated Encoder, respectively, while Table~\ref{tab:perclass} summarizes the corresponding per class accuracies. The total number of test samples per class were 63 (Glioma), 163 (Meningioma), 163 (Pituitary), and 170 (No Tumor).

Averaged over all 559 test samples, this corresponds to an overall accuracy of 91.4\% for the Federated model versus 90.0\% for the local model. The largest single class difference is observed for the Pituitary class, where the Federated Encoder outperforms the local model by nearly 9 percentage points, since the local confusion matrix (Fig.~\ref{fig:confusion_fed}) shows 23 Pituitary cases being misclassified as No Tumor, a confusion largely resolved once the model is trained across the federated clients.

\begin{table}[!t]
\centering
\scriptsize
\setlength{\tabcolsep}{4pt}
\renewcommand{\arraystretch}{1.2}
\caption{Per class performance: Federated Encoder vs. local model.}
\label{tab:perclass}
\begin{tabularx}{\linewidth}{>{\raggedright\arraybackslash}X >{\centering\arraybackslash}X >{\centering\arraybackslash}X >{\centering\arraybackslash}X >{\centering\arraybackslash}X}
\toprule
\textbf{Class} & \textbf{Fed (C/T)} & \textbf{Acc.} & \textbf{Local (C/T)} & \textbf{Acc.} \\
\midrule
Glioma     & 54/63   & 85.7\% & 56/63   & 88.9\% \\
Meningioma & 148/163 & 90.8\% & 145/163 & 88.9\% \\
Pituitary  & 154/163 & 94.5\% & 137/163 & 85.6\% \\
No Tumor   & 155/170 & 91.2\% & 165/170 & 94.8\% \\
\midrule
\textbf{Overall} & \textbf{511/559} & \textbf{91.4\%} & \textbf{503/559} & \textbf{90.0\%} \\
\bottomrule
\end{tabularx}
\end{table}

\subsubsection{Precision, Recall, and F1 Score Comparison}
\begin{figure}[t]
\centering
\captionsetup[subfloat]{labelformat=parens,labelsep=space,font=footnotesize}
\subfloat[Precision.]{\includegraphics[width=0.32\linewidth]{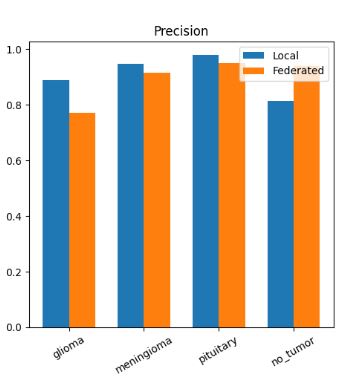}\label{fig:call}}\hfill
\subfloat[Recall.]{\includegraphics[width=0.32\linewidth]{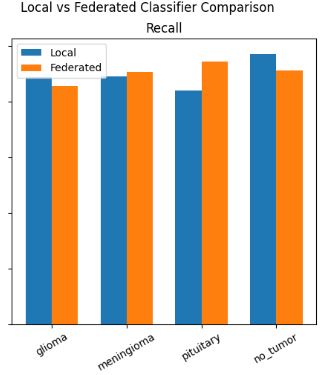}\label{fig:call2}}\hfill
\subfloat[F1 score.]{\includegraphics[width=0.32\linewidth]{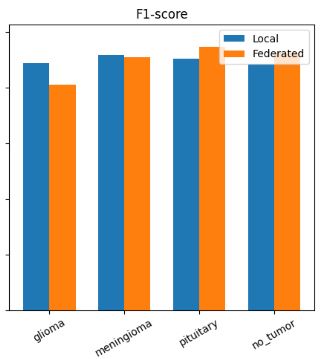}\label{fig:call3}}
\caption{Precision, recall, and F1 score comparison between the local and Federated classifiers across tumor classes.}
\label{fig:prf1_grid}
\end{figure}

Table~\ref{tab:prf1} details the per class Precision, Recall, and F1 score for the local classifier, the same model discussed in Section V C 1, compared against the Federated classifier. Figs.~\ref{fig:call} to \ref{fig:call3} present the corresponding bar plot comparisons.

\begin{table}[!t]
\centering
\scriptsize
\setlength{\tabcolsep}{3pt}
\renewcommand{\arraystretch}{1.2}
\caption{Precision, Recall, and F1 score: local classifier vs. Federated classifier.}
\label{tab:prf1}
\begin{tabularx}{\linewidth}{>{\raggedright\arraybackslash}X >{\centering\arraybackslash}X >{\centering\arraybackslash}X >{\centering\arraybackslash}X >{\centering\arraybackslash}X >{\centering\arraybackslash}X >{\centering\arraybackslash}X}
\toprule
\textbf{Class} & \textbf{P(C)} & \textbf{P(F)} & \textbf{R(C)} & \textbf{R(F)} & \textbf{F1(C)} & \textbf{F1(F)} \\
\midrule
Glioma     & 0.89 & 0.77 & 0.89 & 0.86 & 0.89 & 0.81 \\
Meningioma & 0.95 & 0.91 & 0.89 & 0.91 & 0.92 & 0.91 \\
Pituitary  & 0.98 & 0.95 & 0.84 & 0.94 & 0.90 & 0.95 \\
No Tumor   & 0.81 & 0.94 & 0.97 & 0.91 & 0.88 & 0.93 \\
\bottomrule
\end{tabularx}
\end{table}

Table~\ref{tab:prf1} and Fig.~\ref{fig:call} show that Glioma is the one class where the Federated model underperforms the local model, with precision falling from 0.89 to 0.77, a drop of about 12 percentage points, and recall falling slightly from 0.89 to 0.86, together bringing the Glioma F1 score down from 0.89 to 0.81 as seen in Fig.~\ref{fig:call3}. This drop is real and is reported here without minimizing it. Glioma is also the smallest class in the test set, with only 63 samples, so its metrics are naturally more sensitive to how the class is distributed and aggregated across federated clients, and this concentration effect is the most direct explanation for why the drop appears on this class specifically rather than on the larger classes. Fig.~\ref{fig:call2} shows the Federated model reaching a recall of 0.94 for Pituitary compared to 0.84 for the local model, and Fig.~\ref{fig:call3} shows the Pituitary F1 score improving from 0.90 to 0.95 and the No Tumor F1 score improving from 0.88 to 0.93 under federation, while Meningioma remains essentially unchanged at 0.92 versus 0.91. Taken together, the Federated model gives up precision on one minority class while gaining on two of the remaining three classes, so its overall accuracy of 91.4\% remains close to, and in fact slightly above, the 90.0\% achieved by the local model that had direct access to the pooled data.

\subsubsection{ROC Curve and AUC Analysis}
\begin{figure}[t]
\centering
\captionsetup[subfloat]{labelformat=parens,labelsep=space,font=footnotesize}
\subfloat[Glioma (AUC $\approx 0.99$).]{\includegraphics[width=0.48\linewidth]{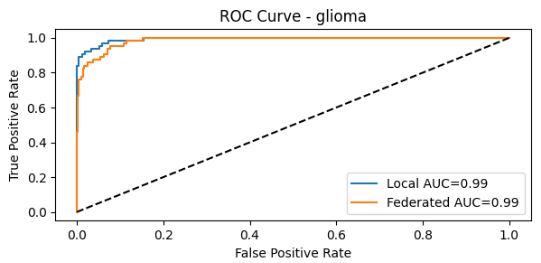}\label{fig:roc1}}\hfill
\subfloat[Meningioma (AUC $\approx 0.98$).]{\includegraphics[width=0.48\linewidth]{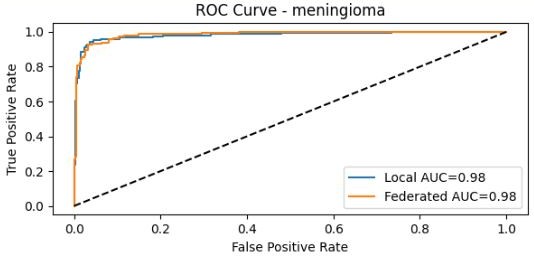}\label{fig:roc2}}\\
\subfloat[Pituitary (AUC $\approx 0.99$).]{\includegraphics[width=0.48\linewidth]{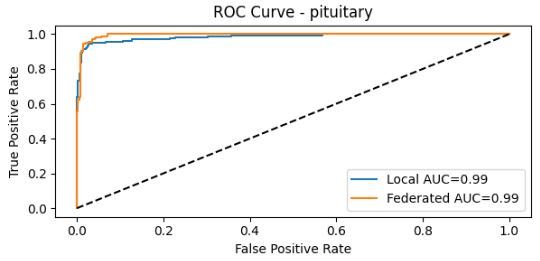}\label{fig:roc3}}\hfill
\subfloat[No Tumor (AUC $\approx 0.98$).]{\includegraphics[width=0.48\linewidth]{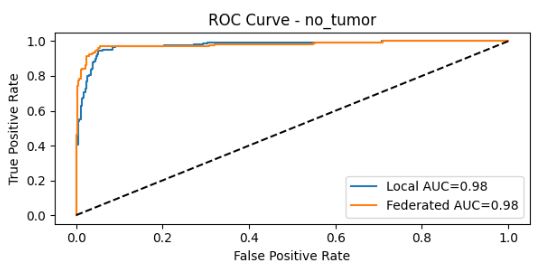}\label{fig:roc4}}
\caption{ROC curves for the local and Federated classifiers across the four tumor classes.}
\label{fig:roc_grid}
\end{figure}

Figs.~\ref{fig:roc1} to \ref{fig:roc4} show the ROC curves for the local and Federated classifiers on the Glioma, Meningioma, Pituitary, and No Tumor classes, respectively. Across all four figures, the two curves nearly overlap, with individual class AUC values of 0.99 (Glioma), 0.98 (Meningioma), 0.99 (Pituitary), and 0.98 (No Tumor), giving an average AUC of approximately 0.985 for both models. This overlap is a useful check on the precision drop discussed above. Even for Glioma, where precision and F1 score fall under federation, the ROC curve and AUC remain essentially unchanged, which indicates that the model ranking ability and overall separability between classes is preserved, and that the drop is concentrated in a specific operating threshold rather than reflecting a general breakdown of the model discriminative power.

\subsection{Model Visualization and Case Study Interpretation}
\begin{figure}[t]
\centering
\includegraphics[width=\linewidth,height=0.80\linewidth,keepaspectratio]{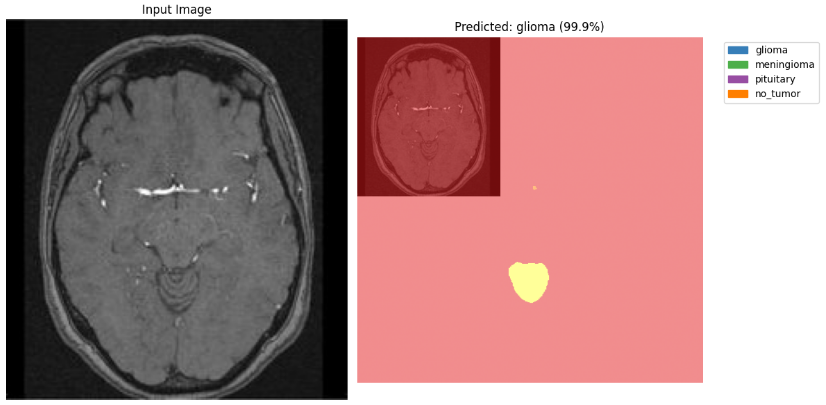}
\caption{Representative case study output: input MRI slice (left) and predicted segmentation with classification label and confidence score (right), for a Glioma sample.}
\label{fig:case_study}
\end{figure}

Fig.~\ref{fig:case_study} shows a representative single sample case output, in which an MRI slice is overlaid with the predicted segmentation mask and classification label. In this example the model classified the sample as Glioma, and the output text reported a confidence score of 98.59\% for this specific sample. This is the sample level figure that has, in earlier drafts, been imprecisely rounded to 99.9\%, and it should not be read as the model overall classification accuracy, which is reported separately in Table~\ref{tab:perclass} as 91.4\% overall and 94.5\% peak per class. The segmentation overlay localizes an anatomically plausible lesion, and the smooth, spatially coherent boundary of the predicted region reflects the regularizing effect of the physics informed loss term.

\begin{figure*}[t]
\centering
\captionsetup[subfloat]{labelformat=parens,labelsep=space,font=footnotesize}
\setlength{\abovecaptionskip}{2pt}
\setlength{\belowcaptionskip}{2pt}
\setlength{\tabcolsep}{0pt}
\renewcommand{\arraystretch}{1}

\subfloat[Glioma (84\%): Accurate segmentation consistent with infiltrative growth.]{
  \includegraphics[width=0.18\textwidth]{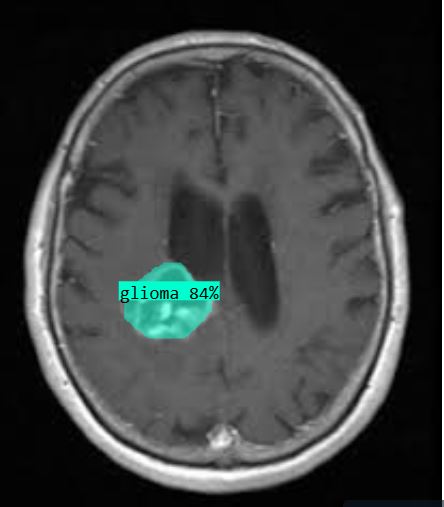}
  \label{fig:glioma_side}}
\hfill
\subfloat[Meningioma (85\%): Precisely located near the dura mater (extra axial tumor), demonstrating spatial context learning.]{
  \includegraphics[width=0.18\textwidth]{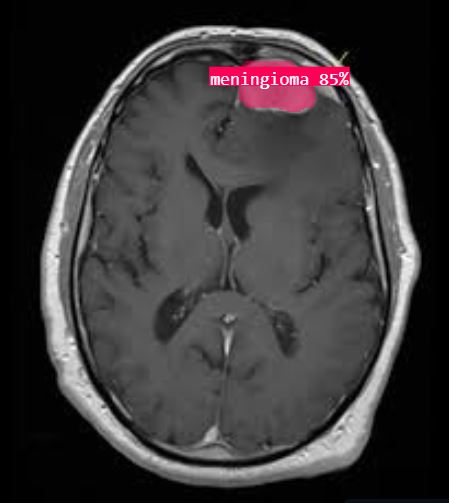}
  \label{fig:meningioma_side}}
\hfill
\subfloat[Pituitary Adenoma (87\%): Correctly situated at the sellar region, confirming learned tissue deformation.]{
  \includegraphics[width=0.18\textwidth]{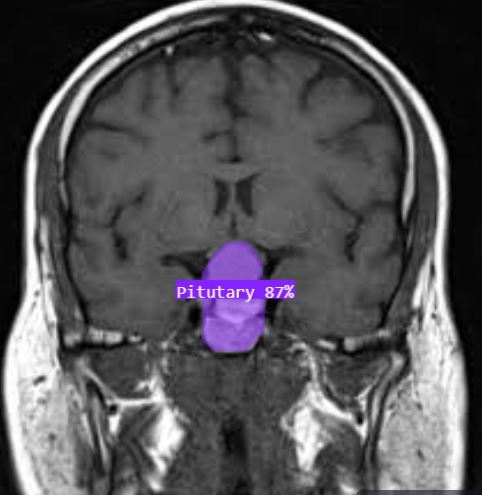}
  \label{fig:pituitary_side}}
\hfill
\subfloat[No Tumor (93\%): Correctly identified as tumor free, minimizing false positives.]{
  \includegraphics[width=0.18\textwidth]{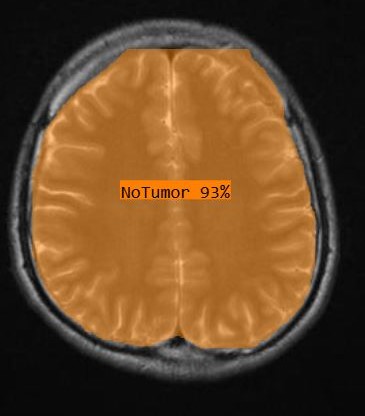}
  \label{fig:notumor_side}}

\caption{Multi class predictions from the Federated PINN model, showing consistent localization and confidence across all four tumor categories.}
\label{fig:multiclass_side}
\end{figure*}

Fig.~\ref{fig:multiclass_side} presents multi class predictions across all four categories. The model localizes Glioma in the parietal region consistent with infiltrative growth (84\% confidence), identifies Meningioma as an extra axial mass adjacent to the dura mater (85\% confidence), situates Pituitary Adenoma correctly within the sellar region (87\% confidence), and correctly labels healthy scans as tumor free (93\% confidence). This consistent, anatomically appropriate localization across all four classes, at confidence levels broadly in line with the 91.4\% overall accuracy reported above, indicates that the federated model generalizes across varying imaging conditions rather than relying on the atypically high single sample confidence shown in Fig.~\ref{fig:case_study}. In summary, the Federated model matches or exceeds the local model on three of the four classes and on overall accuracy, with a single, well isolated precision drop on the smallest class, Glioma, while removing the requirement for raw patient data to leave its originating site.

\section{Novelty and Contribution of the Proposed Hybrid Framework}
\justifying
The proposed Federated Physics Informed Neural Network framework represents a methodological contribution to computational neuro oncology, combining decentralized training with a physically grounded displacement model, and shifting part of the analysis from static image classification alone towards a setting where tumor induced tissue deformation is also modeled explicitly.

\subsection{Innovation and Architectural Novelty}
The primary novelty resides in the pairing of a Physics Informed Neural Network with a Federated Learning protocol within a single pipeline.

\textbf{1) Physically Grounded Displacement Modeling:} Unlike conventional Federated Learning applications that focus solely on fixed image tasks such as detection or segmentation, this framework additionally trains a displacement field model that is constrained by the governing equations of linear elasticity. The core design embeds the Navier Cauchy equilibrium equations directly into the loss function of the PINN, so that the predicted displacement field is not only fit to the boundary condition but is also required to satisfy the underlying stress equilibrium relation for tissue modeled as brain or tumor material.

\textbf{2) Physics Informed Loss Optimization:} The use of a physics informed residual addresses a limitation of purely data driven displacement models, which have no mechanism to enforce consistency with continuum mechanics. By combining a boundary term with the elasticity residual, the PINN is guided towards solutions that remain smooth and divergence free across the domain. The training dynamics support this: the PINN loss curve converges rapidly towards a near zero value within the first hundred epochs, indicating that both the boundary condition and the physics residual are jointly satisfied.

\textbf{3) Privacy Preserving Collaboration:} The adoption of Federated Learning establishes a training protocol in which multiple simulated clinical sites contribute to a shared segmentation and classification model without any raw patient image leaving its site of origin. Local models are trained individually at each client, and only the resulting model weights are shared with a central server, which aggregates them through weighted averaging. This addresses the practical constraint imposed by regulations such as GDPR and HIPAA, since no raw imaging data is transmitted at any stage, although the present implementation relies on standard weight sharing rather than any additional cryptographic or secure aggregation mechanism, and does not itself constitute a formal privacy guarantee.

\subsection{Empirical Performance and Generalizability}
In addition to its architectural design, the proposed framework was evaluated empirically to assess whether decentralized training comes at a meaningful cost to classification performance, and the results are reported here in full, including the cases where the federated model does not improve on the baseline.

\textbf{Overall Classification Accuracy:} The federated framework achieved an overall classification accuracy of 91.4 percent, compared with 90.0 percent for the non federated baseline trained with direct access to the pooled data. This improvement is modest and reflects the benefit of training across a more diverse, cross site partition of the data rather than any physics related effect, since the classification head is trained independently of the PINN module using a standard Cross Entropy objective on features extracted from the shared encoder.

\textbf{Diagnostic Capability:} The ROC curve analysis yielded an average AUC of approximately 0.985 across all four tumor classes, matching the discriminative capability of the centralized baseline while training was carried out without pooling raw patient data. Individual class AUC values reached as high as 0.99 for Glioma and Pituitary.

\textbf{Class Level Generalization Gains:} Pituitary tumor classification accuracy improved from 85.6 percent under the centralized baseline to 94.5 percent under the federated model, resolving a substantial share of the misclassifications into the No Tumor class observed in the baseline confusion matrix. The No Tumor F1 score also improved from 0.88 to 0.93 under federation, and Meningioma performance remained essentially unchanged between the two settings.

\textbf{Class Level Trade Off:} These gains are not uniform across all classes. Glioma is the one class where the federated model underperforms the baseline, with precision falling from 0.89 to 0.77 and the F1 score falling from 0.89 to 0.81. As discussed in Section IV, Glioma is also the smallest class in the test set, which makes its metrics more sensitive to how samples are distributed across clients. This drop is reported here directly rather than omitted, since presenting only the classes that improve would misrepresent the overall behavior of the federated model. Taken together, the framework yields a small net improvement in overall accuracy while trading precision on the smallest class for gains on two of the remaining three classes, and this trade off, rather than a uniform improvement, is the empirical contribution being reported.

\FloatBarrier
\section{Discussion and Limitations}
\justifying
The Federated PINN framework represents a practical step in computational neuro oncology, combining decentralized training with a physically constrained displacement model derived from linear elasticity. By adopting the Federated Learning paradigm, collaborative training across three simulated clinical sites is enabled without exchanging raw patient data, achieving an average AUC of approximately 0.985 and matching the discriminative capability of the centralized baseline. The most notable class level generalization benefit is the improvement in Pituitary tumor classification from 85.6\% to 94.5\% , resolving a substantial share of the misclassifications into the No Tumor class observed in the centralized model. The PINN component further ensures that the learned displacement field respects the boundary condition and the elasticity residual, as confirmed by the stable near zero training loss described in Section IV, though this displacement modeling is carried out on individual static image slices rather than across any temporal sequence, and the framework does not model tumor growth or tissue change over time. Despite these results, several limitations remain. PINN performance is sensitive to imaging noise, which can propagate into the estimated displacement field and compromise its biomechanical reliability. Incorporating a PDE residual into training introduces additional computational cost relative to a purely data driven baseline, which may limit scalability to larger, real time clinical settings. Physical consistency is imposed through a soft loss penalty rather than a hard constraint, so occasional deviations from exact physical consistency remain possible. Spectral bias in the underlying network can favor low frequency components of the displacement field, which may reduce accuracy near sharp tissue boundaries. The two class material model, in which all tumor pixels share a single stiffer Lame parameter pair rather than a subtype specific, literature calibrated value, is a simplification of the true tissue mechanics and may not capture the mechanical differences between glioma, meningioma, and pituitary tissue. Finally, as noted in Section IV, all reported accuracy figures in this study come from a single training run with a fixed random seed.

\section{Conclusion}
\justifying
This study presents a Federated Physics Informed Neural Network framework for brain tumor analysis, combining privacy preserving collaborative training with a physics constrained model of tissue displacement. By integrating Federated Learning, which allows model training to proceed without centralizing patient scans, with a Physics Informed Neural Network that enforces the governing equations of linear elasticity on the predicted displacement field, the framework couples two capabilities that have largely been studied separately in prior work.The improvement was not uniform; the federated model gave up precision on the Glioma class relative to the baseline while improving on the Pituitary and No Tumor classes, and this trade off is reported here as the actual empirical outcome rather than as a uniform gain. Separately, the PINN component demonstrated that a displacement field satisfying the elasticity residual and boundary condition can be learned from segmented MRI slices, producing smooth and physically consistent output on the individual cases examined in this study.The material parameters used for the elasticity residual are fixed input constants rather than patient specific quantities inferred by the model, and no temporal or longitudinal data was used in training. Likewise, while the federated protocol avoids transmitting raw patient images between clients, the present implementation relies on standard weight averaging without additional cryptographic or secure aggregation mechanisms, and should be understood as a step toward, rather than a complete guarantee of, formal data protection compliance. Within these bounds, the results suggest that a physics constrained displacement model can be trained under a federated protocol without a substantial loss in classification performance.

\section{Future Work}
\justifying
Future work will extend the present framework along several directions. Hard constraint mechanisms will be explored in place of the current soft PDE penalty, and frequency aware architectures will be considered to reduce spectral bias near sharp tissue boundaries. The two class material model will be extended to assign subtype specific, literature calibrated Lame parameters to glioma, meningioma, and pituitary tissue individually. The classification and displacement modeling components, currently trained independently, will be investigated for tighter integration. Evaluation will be extended to multiple training runs with variance and significance testing, alongside explicit reporting of per client data distributions. Finally, incorporating longitudinal imaging and paired Magnetic Resonance Elastography data would allow the framework to model tumor growth over time and supervise displacement directly, moving toward patient specific biomechanical estimation for therapy planning.

\FloatBarrier        
\vspace{-1em}        

\FloatBarrier
\vspace{1.2em}

\section{References}
\vspace{-0.5em}
\begingroup
\small
\setlength{\parskip}{1pt}
\setlength{\itemsep}{0.35em}
\renewcommand{\baselinestretch}{0.95}
\renewcommand{\refname}{}

\makeatletter
\renewcommand{\@biblabel}[1]{\textcolor{blue}{[#1]}}
\makeatother

\endgroup

\end{document}